\documentclass[conference]{IEEEtran}
\IEEEoverridecommandlockouts
\usepackage{cite}
\usepackage{amsmath,amssymb,amsfonts}
\usepackage{algorithmic}
\usepackage{graphicx}
\usepackage{textcomp}
\usepackage{comment}
\usepackage{float}
\usepackage{caption}
\usepackage{url}
\usepackage{subcaption}

\usepackage{xcolor}
\def\BibTeX{{\rm B\kern-.05em{\sc i\kern-.025em b}\kern-.08em
    T\kern-.1667em\lower.7ex\hbox{E}\kern-.125emX}}
\begin{document}
\raggedbottom

\title{Mixture-of-Experts Models in Vision: Routing, Optimization, and Generalization%
\thanks{Code: \protect\url{https://github.com/moe-project-uu/mixture-of-experts-project}}}

\author{
\IEEEauthorblockN{Adam Rokah, Daniel Veress, Caleb Caulk, Sourav Sharan}

\IEEEauthorblockA{Department of Information Technology, Uppsala University, Uppsala, Sweden}

\IEEEauthorblockA{\texttt{\{adam.rokah.6563, daniel.veress.5798, caleb.caulk.5143, sourav.sharan.1411\}@student.uu.se}}

\IEEEauthorblockA{\vspace{0.2em}}

\IEEEauthorblockA{Supervised by: Prashant Singh, Salman Toor}

\IEEEauthorblockA{
Department of Information Technology, Uppsala University, Uppsala, Sweden \\
Scaleout Systems, Uppsala, Sweden
}

\IEEEauthorblockA{\texttt{prashant.singh@it.uu.se, salman@scaleoutsystems.com}}
}

\maketitle

\begin{abstract}
Mixture-of-Experts (MoE) architectures enable conditional computation by routing inputs to multiple expert subnetworks and are often motivated as a mechanism for scaling large language models. In this project, we instead study MoE behavior in an image classification setting, focusing on predictive performance, expert utilization, and generalization. We compare dense, SoftMoE, and SparseMoE classifier heads on the CIFAR-10 dataset under comparable model capacity. Both MoE variants achieve slightly higher validation accuracy than the dense baseline while maintaining balanced expert utilization through regularization, avoiding expert collapse. To analyze generalization, we compute Hessian-based sharpness metrics at convergence, including the largest eigenvalue and trace of the loss Hessian, evaluated on both training and test data. We find that SoftMoE exhibits higher sharpness by these metrics, while Dense and SparseMoE lie in a similar curvature regime, despite all models achieving comparable generalization performance. Complementary loss-surface perturbation analyses reveal qualitative differences in non-local behavior under finite parameter perturbations between dense and MoE models, which help contextualize curvature-based measurements without directly explaining validation accuracy. We further evaluate empirical inference efficiency and show that naively implemented conditional routing does not yield inference speedups on modern hardware at this scale, highlighting the gap between theoretical and realized efficiency in sparse MoE models.
\end{abstract}

\begin{IEEEkeywords}
Mixture of Experts, Generalization, Sharpness
\end{IEEEkeywords}

\section{Introduction}

Recent advances in machine learning have led to substantial performance improvements across domains such as computer vision, speech recognition, and natural language processing \cite{AeRviewOfConvolutionalNeuralNetworksInComputerVision,DeepSpeech2,LargeLanguageModelsASurvey}. Many of these gains have been driven by increased model capacity, training data, and computational resources, particularly in large-scale dense architectures \cite{ScalingLawsForNeuralLanguageModels}. While effective, this trend raises questions about efficiency and how architectural choices influence optimization and generalization beyond raw scale.

Mixture-of-Experts (MoE) architectures offer an alternative design by introducing conditional computation, routing inputs to a subset of specialized expert subnetworks rather than activating all parameters simultaneously \cite{shazeer2017outrageously}. In large language models, MoEs are primarily motivated by efficiency and scalability \cite{SwitchTransformers,deepseekv3technicalreport}. However, comparatively little work has examined how MoE design choices affect optimization behavior and generalization in smaller, controlled settings where efficiency is not the primary experimental focus.

In this work, we study Mixture-of-Experts models in an image classification context, deliberately decoupling MoE behavior from large-scale efficiency considerations. Using the CIFAR-10 dataset, we compare dense classifier heads with SoftMoE and SparseMoE variants built on a shared ResNet-18 backbone and matched to comparable parameter budgets. This controlled setup allows us to isolate the effects of expert routing, load balancing, and sparsity on predictive performance and training dynamics.

Beyond validation accuracy, we analyze generalization-related behavior through the geometry of the loss landscape. Specifically, we compute Hessian-based sharpness metrics at convergence and complement these local measures with loss-surface perturbation analyses. While such metrics are often used as diagnostics of local stability, their relationship to predictive performance is known to be nuanced.

Overall, our results show that Mixture-of-Experts models generalize comparably to dense models under matched capacity, despite exhibiting differences in local curvature statistics across architectures. Importantly, Hessian-based sharpness metrics for MoE models lie in a similar regime to those of dense baselines, suggesting that the observed performance is not driven by anomalous curvature or instability in our experimental setup. Additional loss-surface perturbation analyses reveal differences in loss behavior under finite parameter perturbations associated with routing, which provide contextual insight into optimization geometry but do not directly account for validation accuracy.

\section{Methodology} \label{Methodology}

\subsection{Mixture-of-Experts Overview}

A Mixture-of-Experts (MoE) model replaces a single classifier with multiple expert subnetworks and a gating mechanism that determines how each input is processed. In our image classification setting, each input image is first mapped to a feature representation by a shared convolutional backbone. This feature vector is then passed to either a dense classifier head or an MoE-based head.

In an MoE head, a gating network assigns routing scores to a set of expert networks, each of which produces class logits from the same feature representation. Depending on the MoE variant, either all experts contribute to the final prediction or only a subset of experts is evaluated per input, enabling expert specialization and, in sparse variants, conditional computation.

\subsection{MoE Architecture and Variants}

The Mixture-of-Experts architecture was originally introduced as a form of conditional computation in which inputs are dynamically routed to specialized subnetworks \cite{jacobsoriginalMoE}. Formally, the output of an MoE layer is
\begin{equation}
y = \sum_{i=1}^{N} G_i(x)\, E_i(x),
\end{equation}
where $E_i(x)$ denotes the output of the $i$-th expert and $G_i(x)$ is the corresponding gating weight.

\begin{figure}[t]
    \centering
    \includegraphics[width=.9\columnwidth]{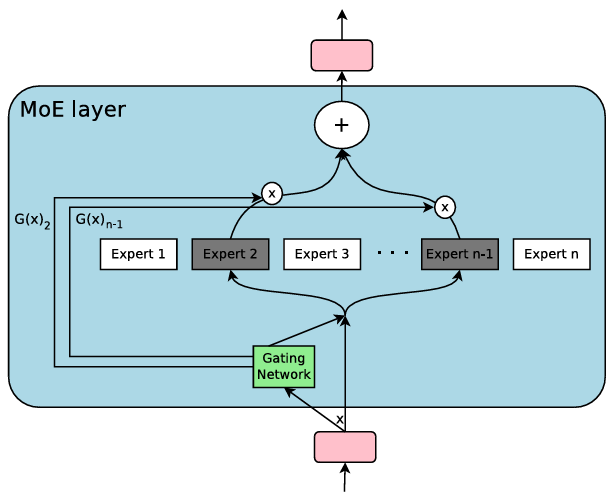}
    \caption{Conceptual illustration of a Mixture-of-Experts architecture, where a gating network routes inputs to multiple expert subnetworks \cite{shazeer2017outrageously}.}
    \label{fig:MoEArchitecture}
\end{figure}

To illustrate this routing mechanism, Fig.~\ref{fig:MoEArchitecture} provides a conceptual visualization of an MoE model, highlighting the interaction between the gating network and expert subnetworks.

We consider three MoE variants. In \textit{SoftMoE}, all experts are evaluated for each input, and their outputs are combined using a softmax over gating logits. In \textit{SparseMoE}, only the top-$k$ experts are selected per input using a Shazeer-style Noisy Top-$k$ gating mechanism \cite{shazeer2017outrageously}, and only selected experts are evaluated in the forward pass. \textit{HardMoE} is the special case $k=1$. We implement both HardMoE ($k=1$) and SparseMoE ($k=2$), but focus our main analysis on $k=2$, which preserves conditional computation while allowing smoother mixture behavior and more stable optimization.

\subsection{Load Balancing}

A key challenge in MoE training is expert collapse, where routing concentrates on a small subset of experts. To encourage balanced expert utilization, we apply load-balancing regularization across all MoE variants. For SoftMoE, we include a batch-level regularization term given by the Kullback--Leibler (KL) divergence between the average routing distribution and a uniform prior, which enforces balanced usage in expectation while allowing per-input specialization. For SparseMoE and HardMoE, we incorporate auxiliary importance and load losses following Shazeer et al.~\cite{shazeer2017outrageously}, together with injected gating noise to stabilize early training, applied after normalizing gating logits to ensure that the noise operates on a consistent scale. During inference, routing is deterministic, with gating noise disabled, ensuring stable and reproducible evaluation.

\subsection{Experimental Setup}

All experiments are conducted on the CIFAR-10 dataset\cite{krizhevsky2009learning}. It consists of 60,000 32×32 color images distributed across 10 distinct classes, with 6,000 images per class. A shared ResNet-18 backbone is used with a CIFAR-specific stem. The backbone maps each input image $x \in \mathbb{R}^{3 \times 32 \times 32}$ to a feature vector $h \in \mathbb{R}^{512}$. This feature representation serves as the input to all classifier heads.

\begin{figure*}[t]
    \centering
    \includegraphics[width=\textwidth]{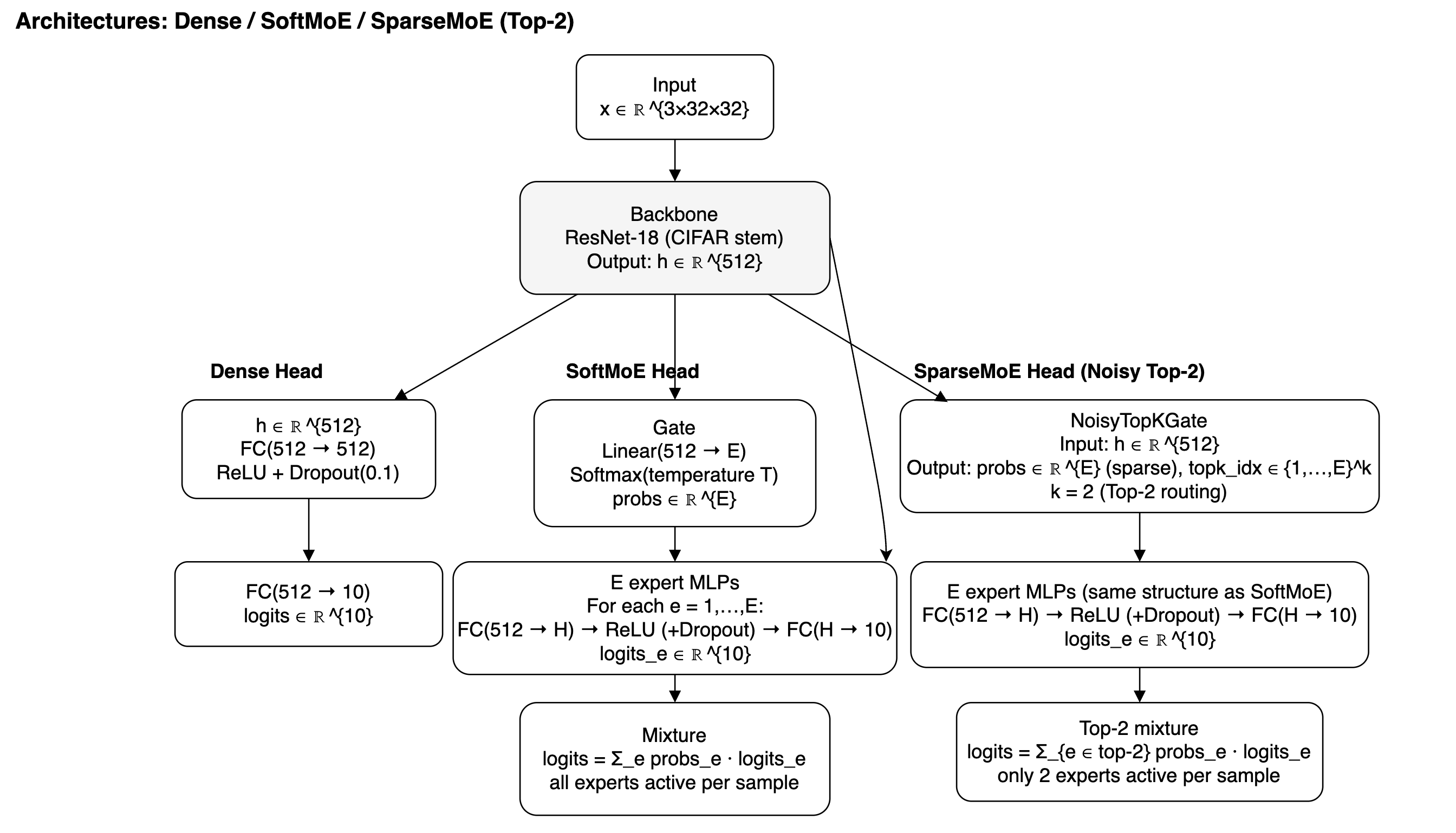}
    \caption{Architectures used in this study. A shared ResNet-18 backbone produces a 512-dimensional feature vector, which is processed by either a dense classifier head, a SoftMoE head with all experts active, or a SparseMoE head using Noisy Top-$2$ routing.}
    \label{fig:architecture}
\end{figure*}

In the dense baseline, $h$ is processed by a single multilayer perceptron. In the MoE variants, $h$ is passed to a gating network and multiple expert multilayer perceptrons operating on the same feature representation. Dense, SoftMoE, and SparseMoE models are constructed with comparable parameter budgets by adjusting the number of experts and the hidden dimensionality of each expert. Fig.~\ref{fig:architecture} summarizes the architectures used in our experiments.

All models are trained using stochastic gradient descent with momentum and weight decay under identical optimization settings. Validation accuracy is used for model selection, and final performance is evaluated on a held-out test set.

\subsection{Generalization and Loss Landscape Analysis}

\begin{figure}[H]
    \centering
    \includegraphics[width=.9\columnwidth]{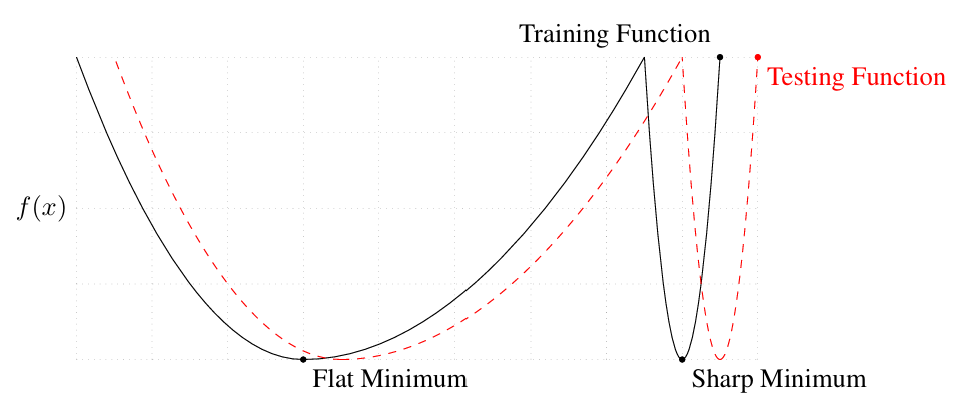}
    \caption{Illustration of flat versus sharp minima in the loss landscape, commonly used to motivate curvature-based generalization diagnostics \cite{onlargebatchtraining}.}
    \label{fig:generalization}
\end{figure}

To analyze generalization-related behavior, we study the geometry of the loss landscape at convergence. One commonly used diagnostic is sharpness, which characterizes local curvature of the loss surface around a solution. Intuitively, flatter minima are less sensitive to small parameter perturbations, while sharper minima exhibit higher sensitivity. Fig.~\ref{fig:generalization} provides a conceptual illustration of this distinction.

We compute Hessian-based sharpness metrics at conver-
gence, including the largest eigenvalue and the trace of the
loss Hessian, using stochastic Hessian–vector products. These
quantities are treated as diagnostics of local curvature and
stability rather than direct predictors of validation accuracy.
To complement these local measures, we also evaluate the
loss under perturbations along dominant Hessian eigenvector
directions, which is particularly relevant in MoE models where
routing behavior can induce loss changes that are not captured
by purely local second-order curvature.

\section{Models} \label{Models}

Differences between models arise solely from the classifier head, allowing for controlled architectural comparisons under matched backbone and training conditions. We consider a dense baseline head and two Mixture-of-Experts variants, SoftMoE and SparseMoE, which differ in their routing mechanisms, expert activation patterns, and load-balancing regularization. All classifier heads are constructed to have comparable parameter budgets by adjusting the number of experts and per-expert hidden dimensions.

In the dense baseline, the 512-dimensional feature vector is passed to a single multilayer perceptron (MLP) to produce class logits. In the MoE variants, the same feature vector is passed to a gating network and a set of expert MLPs, whose outputs are combined according to the routing mechanism described in Section~\ref{Methodology}.

To ensure fair comparisons, Dense, SoftMoE, and SparseMoE models are constructed with comparable parameter budgets by adjusting the number of experts and the hidden dimensionality of each expert network. SparseMoE uses Noisy Top-$k$ routing with $k=2$ for the main experiments. HardMoE ($k=1$) is implemented and verified for stability and accuracy but is not included in the primary generalization analysis.

Table~\ref{tab:models} summarizes the model configurations used in our experiments.

\begin{table}[H]
\centering
\caption{Model configurations used in the experiments. All models share the same ResNet-18 backbone and are matched to comparable parameter budgets.}
\label{tab:models}
\begin{tabular}{lccc}
\hline
Model & Experts (width) & Active & Load balance \\
\hline
Dense     & -- (512) & -- & -- \\
SoftMoE   & 8 (64)   & All & KL to uniform \\
SparseMoE & 8 (64)   & 2   & Importance + load + noise \\
\hline
\end{tabular}
\end{table}

\section{Results} \label{Results}

We now present empirical results comparing Dense, SoftMoE, and SparseMoE
classifier heads under matched model capacity. We focus on three aspects:
(i) predictive performance, (ii) expert utilization and specialization,
and (iii) generalization-related properties as reflected by local curvature
of the loss landscape.

\subsection{Predictive Performance}

Table~\ref{tab:accuracy_results} summarizes predictive performance across
architectures. All models achieve perfect training accuracy, indicating
sufficient capacity to fit the CIFAR-10 training set. Both MoE variants
slightly outperform the dense baseline on validation accuracy, with gains of
approximately $0.3$--$0.4\%$.

We additionally report the epoch-to-threshold (ETT), defined as the epoch at
which maximum training accuracy ($M_A$) or validation accuracy ($V_A$) is
first achieved. MoE models generally reach peak validation performance later
than the dense baseline, reflecting longer but stable optimization dynamics
induced by routing and load-balancing regularization.

\begin{table}[htbp]
\centering
\caption{Predictive performance across architectures.}
\label{tab:accuracy_results}
\begin{tabular}{lcccc}
\hline
Model & $M_A$ (\%) & ETT($M_A$) & $V_A$ (\%) & ETT($V_A$) \\
\hline
Dense      & 100.00 & 37 & 87.86 & 35 \\
SoftMoE   & 100.00 & 41 & 88.24 & 48 \\
SparseMoE & 100.00 & 44 & 88.16 & 46 \\
\hline
\end{tabular}
\end{table}

\subsection{Expert Utilization and Specialization}

We next analyze how experts are utilized during training and whether
specialization emerges. Fig.~\ref{fig:util_group}(a) and Fig.~\ref{fig:util_group}(b)
show the mean routing probability per expert over training epochs for SoftMoE
and SparseMoE, respectively. In both cases, routing distributions converge
towards balanced utilization, indicating that the applied load-balancing
regularization successfully prevents expert collapse.

{\setlength{\intextsep}{4pt}%
 \setlength{\textfloatsep}{4pt}%
 \setlength{\floatsep}{4pt}%
 \begin{figure}[H]
 \centering
 \begin{subfigure}[t]{.48\textwidth}
     \centering
     \includegraphics[width=\linewidth]{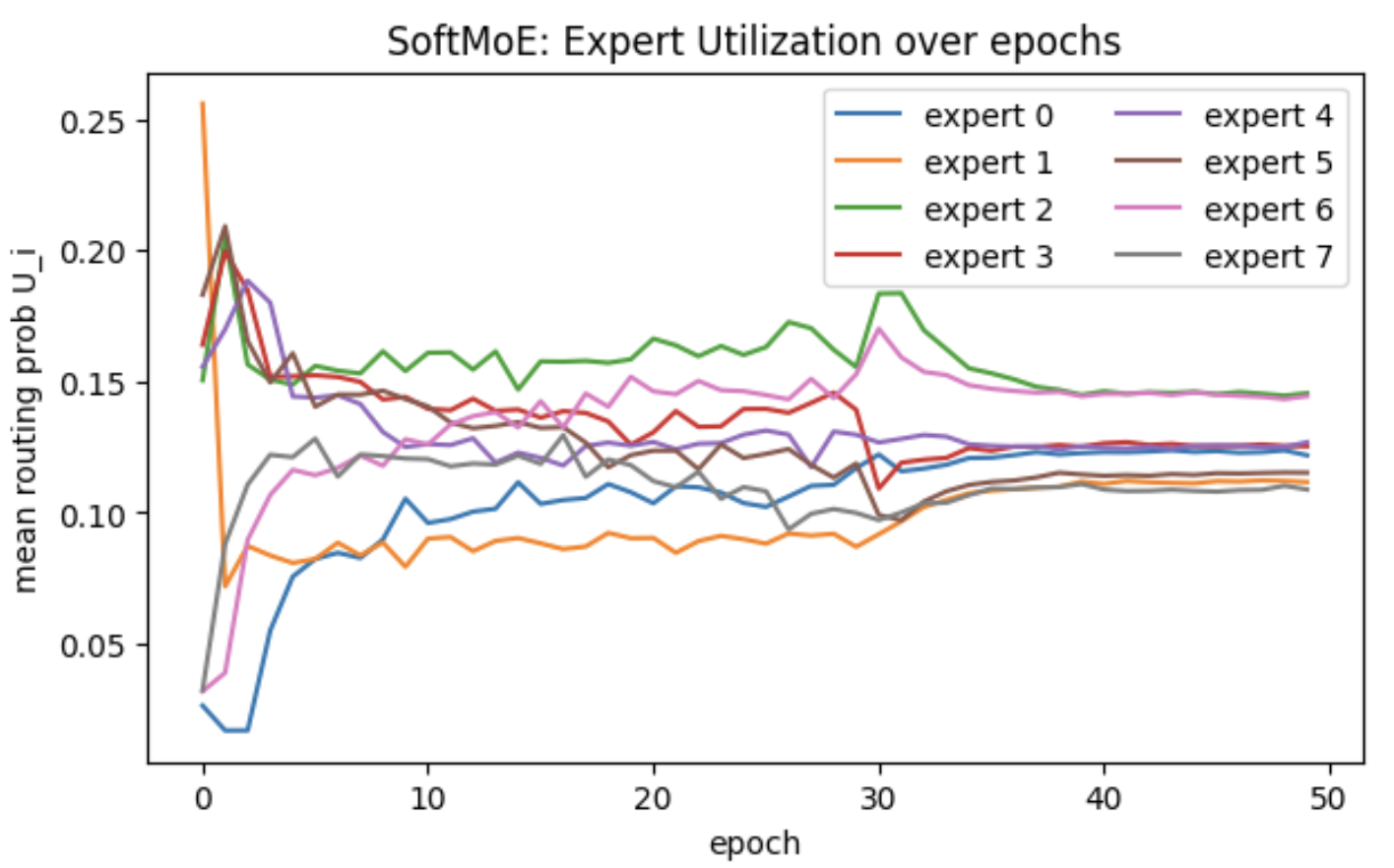}
     \caption{SoftMoE utilization.}
     \label{fig:util_soft}
 \end{subfigure}
 \hfill
 \begin{subfigure}[t]{.48\textwidth}
     \centering
     \hspace{-8pt}
     \includegraphics[width=\linewidth]{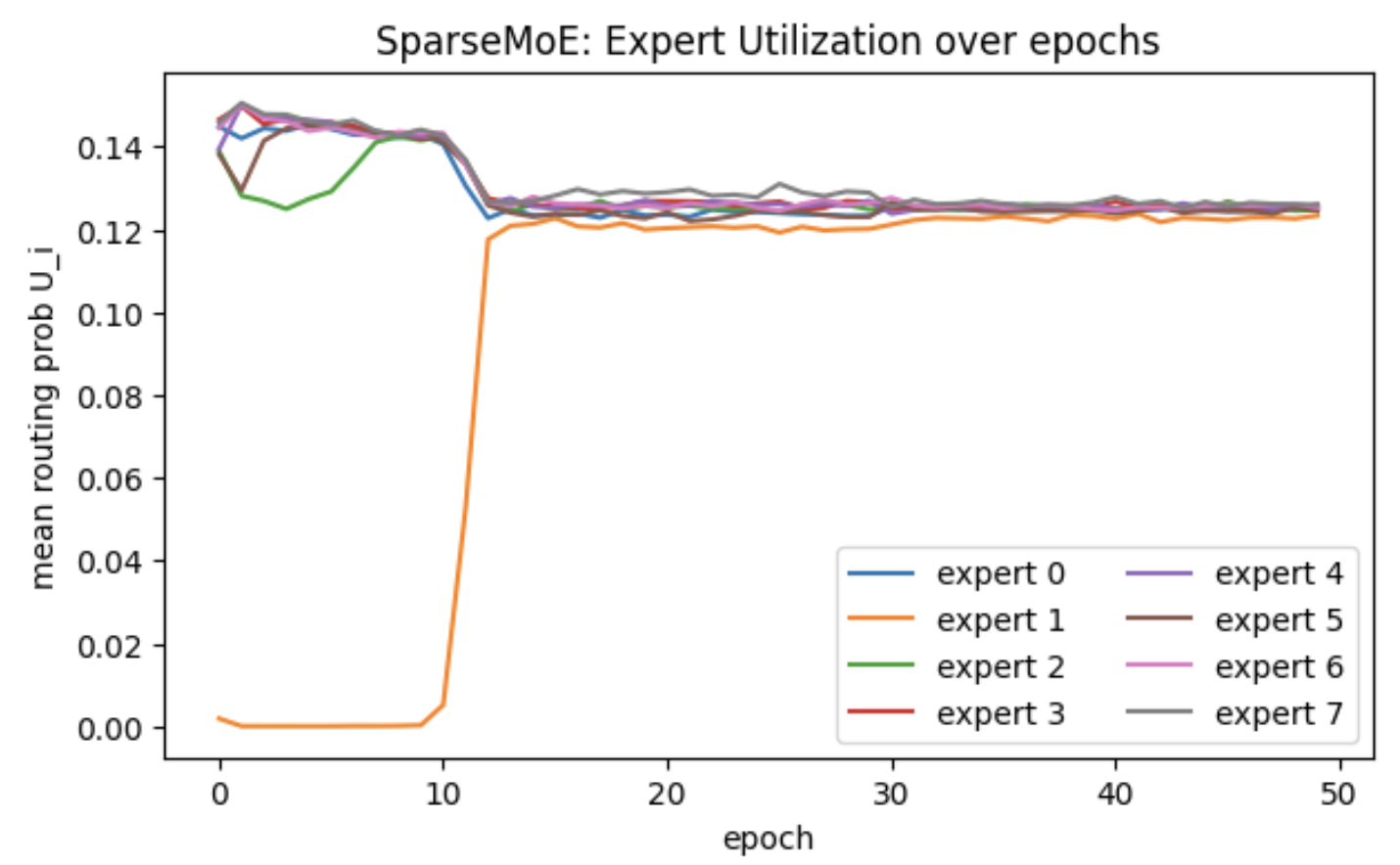}
     \caption{SparseMoE (Top-2) utilization.}
     \label{fig:util_sparse}
 \end{subfigure}
 \caption{Expert utilization over training.}
 \label{fig:util_group}
 \end{figure}
}

To examine specialization, Fig.~\ref{fig:heat_group}(a) and Fig.~\ref{fig:heat_group}(b)
visualize class-conditional expert routing probabilities at convergence. Both MoE variants exhibit clear specialization
patterns: most experts are strongly associated with one or two classes, while
one or two experts contribute more broadly across classes. This behavior
motivates the emergence of ``shared'' experts observed in large-scale MoE
models and suggests that specialization arises naturally even in this
moderate-scale setting.

{\setlength{\intextsep}{4pt}%
 \setlength{\textfloatsep}{4pt}%
 \setlength{\floatsep}{4pt}%
 \begin{figure}[H]
 \centering
 \hspace{-18pt}
 \begin{subfigure}[t]{\columnwidth}
     \centering
     \includegraphics[width=1.03\linewidth]{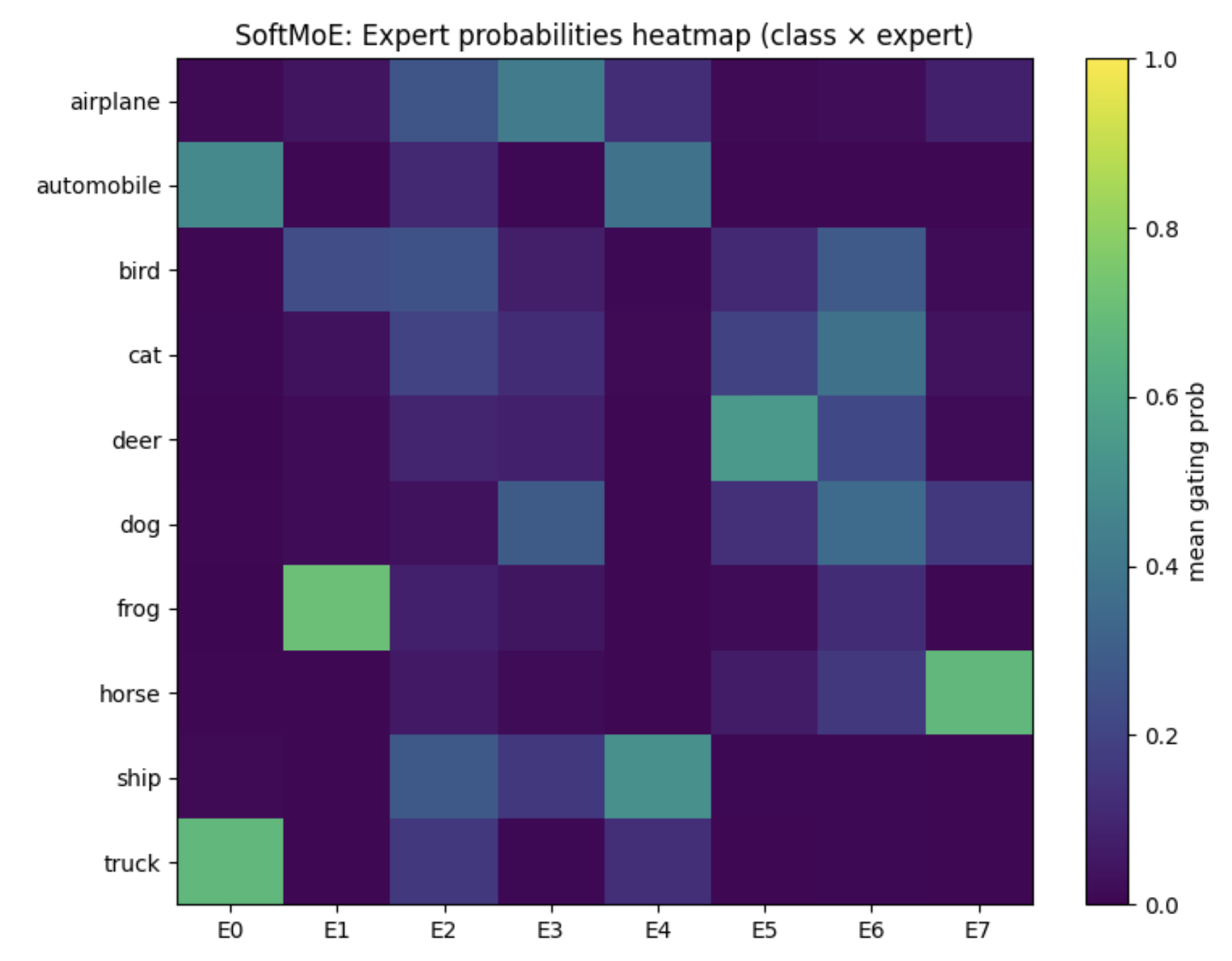}
     \caption{SoftMoE: class--expert routing probabilities at convergence.}
     \label{fig:heat_soft}
 \end{subfigure}

 \vspace{0.3em}

 \begin{subfigure}[t]{\columnwidth}
     \centering
     \includegraphics[width=\linewidth]{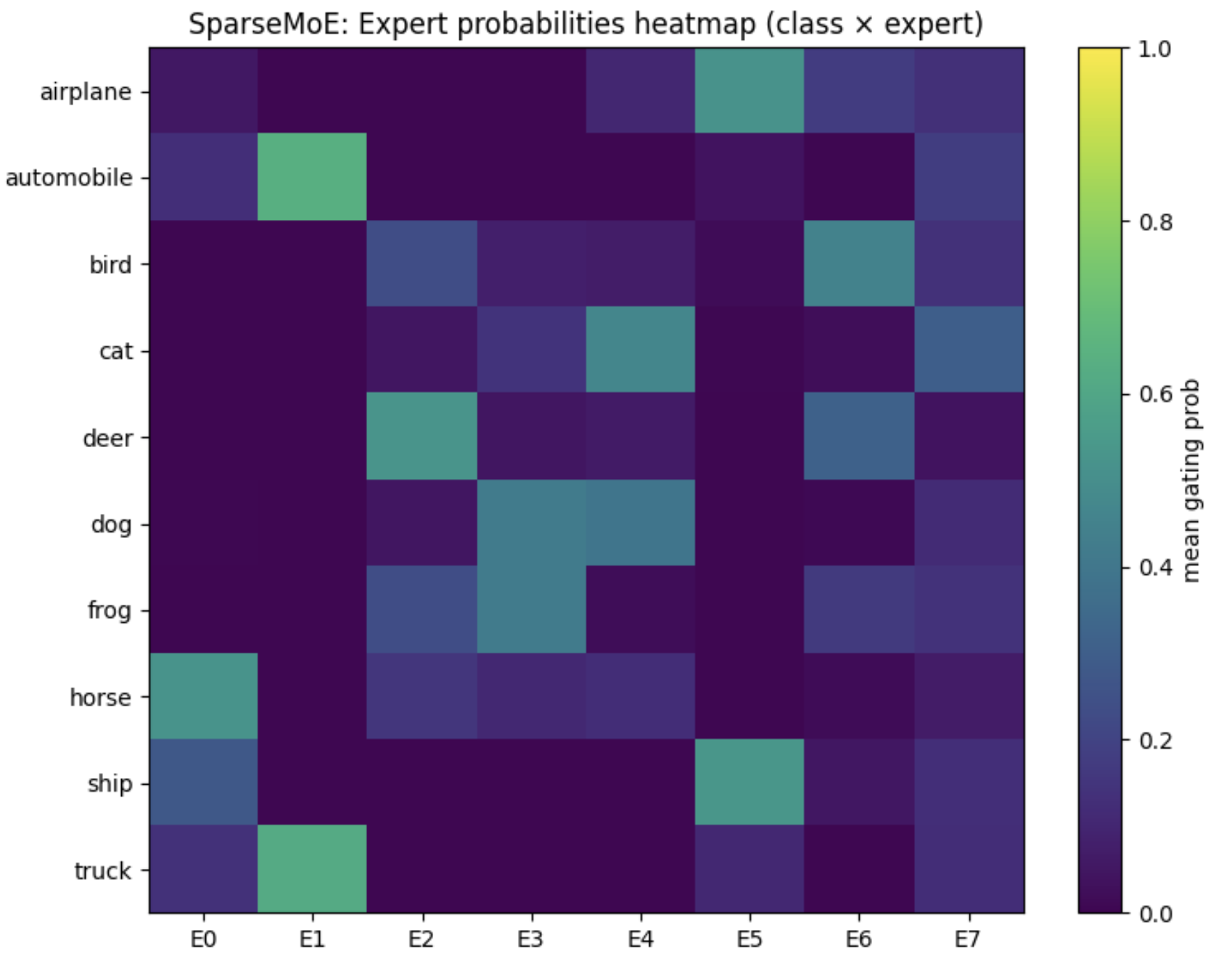}
     \caption{SparseMoE (Top-2): class--expert routing probabilities at convergence.}
     \label{fig:heat_sparse}
 \end{subfigure}
 \caption{Class-conditioned expert specialization at convergence. Both variants exhibit structured specialization, with several experts strongly associated with specific classes and some acting more broadly.}
 \label{fig:heat_group}
 \end{figure}
}

\subsection{Loss Landscape Geometry and Generalization}

We now turn to generalization-related properties through Hessian-based
curvature analysis. Table~\ref{tab:hessian_values} reports the largest Hessian
eigenvalue $\lambda_{\max}$ and Hessian trace evaluated at convergence on both
training and test data.

\begin{table}[H]
\centering
\caption{Hessian-based curvature statistics at convergence.}
\label{tab:hessian_values}
\begin{tabular}{lcccc}
\hline
Model & $\lambda_{\max}^{\text{train}}$ & $\mathrm{Tr}(H)^{\text{train}}$ & $\lambda_{\max}^{\text{test}}$ & $\mathrm{Tr}(H)^{\text{test}}$ \\
\hline
Dense      & 2.38 & 22.75 & 63.87 & 1180.48 \\
SoftMoE    & 1.10 & 19.86 & 92.22 & 1455.34 \\
SparseMoE  & 0.72 & 15.23 & 68.71 & 1148.18 \\
\hline
\end{tabular}
\end{table}

{
\setlength{\intextsep}{2pt}%
\setlength{\textfloatsep}{2pt}%
\setlength{\floatsep}{2pt}%
\captionsetup{skip=2pt}%
\begin{figure}[H]
\centering
\begin{subfigure}[t]{\columnwidth}
    \centering
    \vspace{2pt}
    \includegraphics[width=\linewidth]{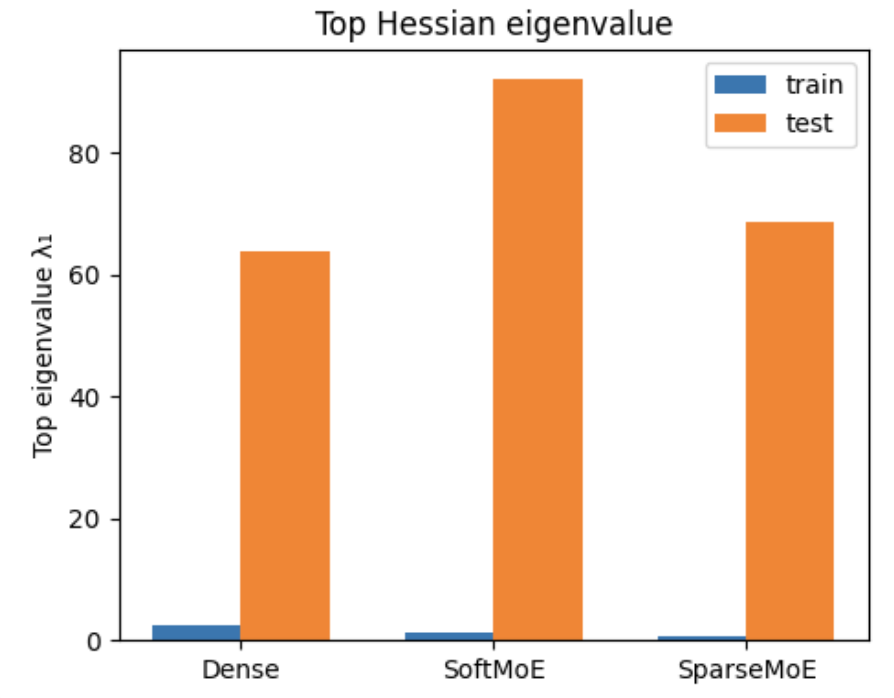}
    \caption{$\lambda_{\max}$}
    \label{fig:hess_lmax}
\end{subfigure}

\vspace{0.2em}

\begin{subfigure}[t]{\columnwidth}
    \centering
    \hspace*{-7pt} 
    \includegraphics[width=1.02\linewidth]{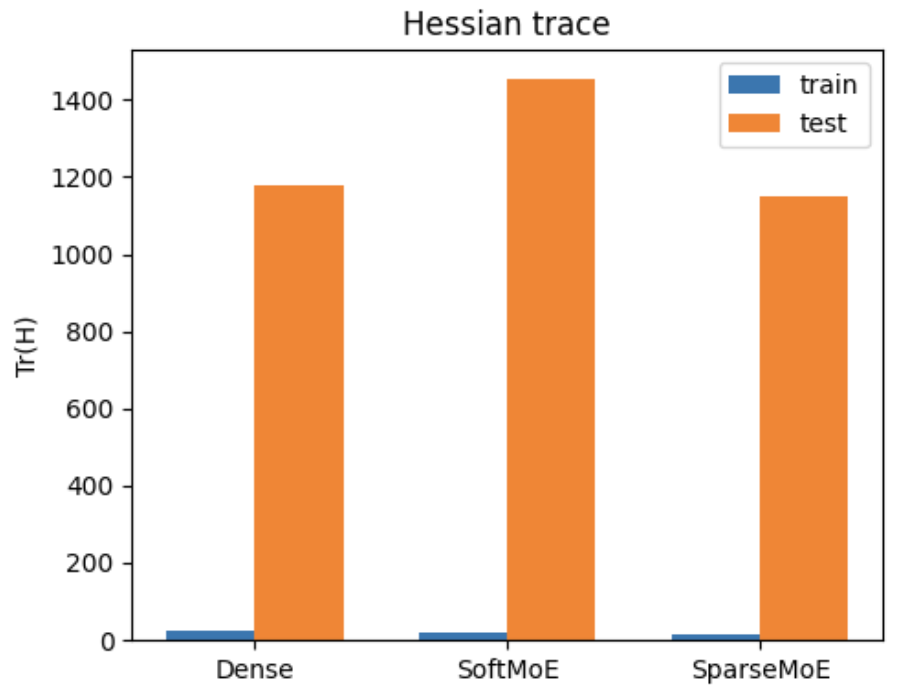}
    \caption{$\mathrm{Tr}(H)$}
    \label{fig:hess_trace}
\end{subfigure}
\caption{Hessian-based curvature statistics at convergence.}
\label{fig:hessian_stats}
\end{figure}
}

SoftMoE exhibits the largest curvature under both test-set metrics, while
Dense and SparseMoE lie in a similar sharpness regime. Importantly, higher
curvature in SoftMoE does not correspond to degraded validation performance,
reinforcing that local sharpness alone is not a reliable predictor of accuracy.

\begin{figure}[H]
\centering
\includegraphics[width=\columnwidth]{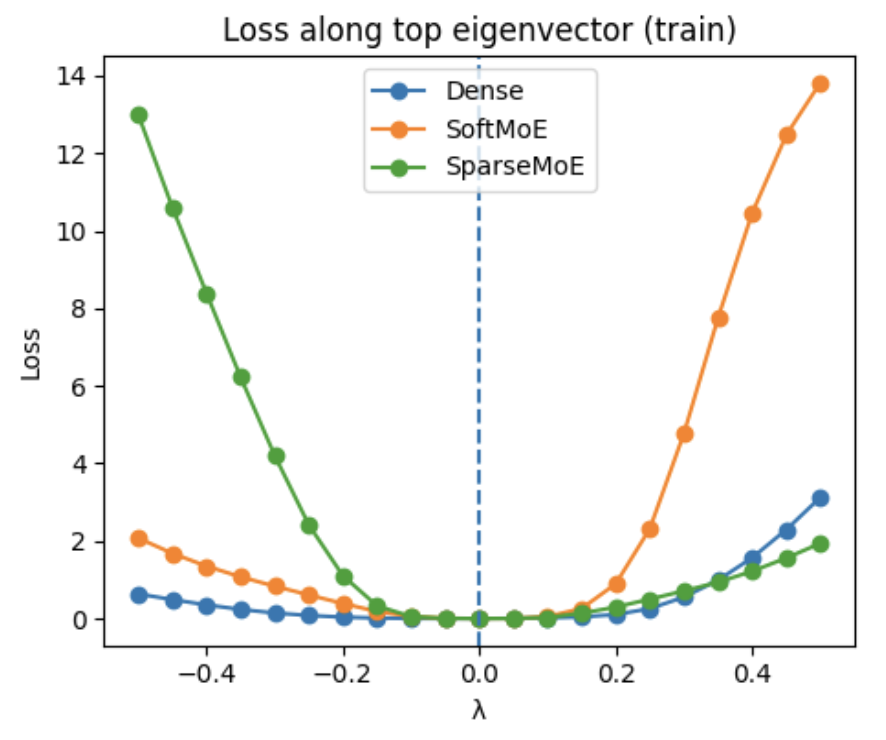}
\caption{Loss along the dominant Hessian eigenvector (training set). Increased sensitivity in MoE models reflects routing changes and gating dynamics rather than purely local curvature.}
\label{fig:hessian_perturb}
\end{figure}

To complement these local metrics, Fig.~\ref{fig:hessian_perturb} shows the loss
evaluated along the dominant Hessian eigenvector. MoE models display sharper
loss growth under larger perturbations compared to the dense baseline. This
behavior reflects the sensitivity of routing decisions: small parameter
changes can redirect inputs to different experts, leading to abrupt changes
in the loss. As a result, these broader perturbation curves primarily capture
global routing effects rather than local flatness near the solution.

Overall, Dense and SparseMoE solutions occupy comparable local curvature
regimes, indicating that the observed MoE performance gains are not an
artifact of unusually flat or unstable solutions. Instead, they reflect
robust behavior under matched capacity, with routing-induced curvature effects
requiring careful interpretation.

\subsection{Theoretical Inference Efficiency}

A primary motivation for Mixture-of-Experts architectures is the potential
for improved inference efficiency through conditional computation. In a dense
classifier head, all parameters contribute to every forward pass. In contrast,
SparseMoE activates only a subset of experts per input, reducing the number of
expert networks evaluated in principle.

In our SparseMoE configuration with $k=2$ active experts out of eight, the
theoretical expert computation per input is reduced by a factor of four
relative to SoftMoE, which evaluates all experts. Ignoring routing overhead,
this suggests a proportional reduction in floating-point operations within
the expert layers. SoftMoE, by contrast, performs full expert evaluation and
thus does not reduce theoretical compute relative to the dense baseline.

\subsection{Empirical Inference Efficiency}

We evaluate empirical inference efficiency by measuring wall-clock latency,
throughput, and peak memory usage during inference on both GPU and CPU. All
measurements are performed with models in evaluation mode and without gradient
tracking. GPU benchmarks are conducted on an NVIDIA A100 using batch size
256, while CPU benchmarks use batch size 32.

\begin{table}[H]
\centering
\caption{Empirical inference efficiency across architectures.}
\label{tab:efficiency}
\begin{subtable}[t]{\columnwidth}
\centering
\caption{GPU inference (NVIDIA A100, batch size 256).}
\label{tab:efficiency_gpu}
\begin{tabular}{lcccc}
\hline
Model & Params (M) & Peak mem (MB) & ms/batch & img/s \\
\hline
Dense      & 11.44 & 421.4 & 9.14  & 27997 \\
SoftMoE    & 11.44 & 421.4 & 9.29  & 27542 \\
SparseMoE  & 11.44 & 421.4 & 13.86 & 18469 \\
\hline
\end{tabular}
\end{subtable}

\vspace{0.6em}

\begin{subtable}[t]{\columnwidth}
\centering
\caption{CPU inference (batch size 32).}
\label{tab:efficiency_cpu}
\begin{tabular}{lccc}
\hline
Model & ms/batch & img/s \\
\hline
Dense      & 70.45 & 454 \\
SoftMoE    & 70.79 & 452 \\
SparseMoE  & 73.01 & 438 \\
\hline
\end{tabular}
\end{subtable}
\end{table}

Table~\ref{tab:efficiency} summarizes empirical inference efficiency across
architectures. Despite activating fewer experts per input, SparseMoE exhibits
higher inference latency and lower throughput than both the dense baseline and
SoftMoE on GPU, a trend that persists on CPU.

This behavior arises because the potential computational savings from skipping
inactive experts are offset by the overhead introduced by conditional routing.
Each input requires computing routing scores, selecting the top-$k$ experts, and
assembling the final prediction from multiple expert outputs. These additional
operations introduce extra control flow and data movement that are absent in
dense models, resulting in less regular execution that is harder to efficiently
optimize on modern hardware. At the batch sizes considered here, these overheads
cannot be amortized across enough inputs to become negligible, preventing
sparsity from yielding practical speedups.

SoftMoE, while evaluating all experts, benefits from fully dense and vectorized
execution across experts, leading to inference performance comparable to the
dense baseline. Peak GPU memory usage is nearly identical across all
architectures, reflecting similar parameter sizes and activation footprints
during inference.

Overall, these results highlight a gap between theoretical and realized
efficiency in sparse MoE models. While conditional computation reduces
theoretical expert FLOPs, translating these reductions into practical inference
speedups requires careful system-level optimization beyond naively implemented conditional routing.

Finally, we note that the efficiency advantages of expert-based
architectures primarily arise in regimes where specialization
reduces representational interference. In settings with high
data or task heterogeneity, dense models must increase the
capacity of a shared representation to accommodate competing
features, increasing computation for every input. MoE architectures instead allow representational capacity to scale selectively through specialized experts, enabling improved predictive performance without proportional increases in per-input computation.

In this study, we consider a relatively homogeneous vision
benchmark and employ a shared backbone with comparatively
thin experts to maintain a controlled setting under matched
model capacity. As a result, effective specialization is limited,
and the overhead introduced by routing and aggregation
dominates any potential savings from sparse expert activation.
Consequently, both performance gains and inference efficiency
differences between dense and expert-based models remain
modest. In more heterogeneous settings, these trade-offs are
expected to shift in favor of MoE-based inference.

\section{Conclusion and Discussion} \label{Conclusion}

In this work, we conducted a controlled empirical study of Dense, SoftMoE, and
SparseMoE architectures in an image classification setting, isolating the
effects of expert routing and load-balancing mechanisms under matched model
capacity. Across all models, we observed perfect training accuracy and
comparable generalization performance, with both MoE variants slightly
outperforming the dense baseline on validation accuracy when expert utilization
is properly regularized.

Through Hessian-based curvature analysis, we found that SoftMoE converges to
solutions with higher local curvature, as measured by the largest Hessian
eigenvalue and trace, while Dense and SparseMoE occupy a similar sharpness
regime. Importantly, these differences in local curvature do not translate into
degraded predictive performance. This result reinforces the view that
sharpness-based diagnostics should be interpreted as indicators of local
stability rather than direct predictors of validation accuracy, a phenomenon
already observed in dense neural networks and here extended to architectures
with input-dependent routing.

Complementary loss-surface perturbation analyses further reveal qualitative
differences between dense and MoE models under finite parameter perturbations.
In MoE architectures, small parameter changes can alter routing decisions,
leading to larger loss variations that reflect global routing sensitivity rather
than local curvature near the solution. These effects provide additional context
for understanding optimization behavior in MoE models but do not by themselves
explain generalization performance.

In addition to predictive and geometric analyses, we evaluated empirical
inference efficiency on both GPU and CPU. Despite reduced expert activation,
SparseMoE did not yield inference speedups relative to dense or SoftMoE models
in our setting. This gap between theoretical and realized efficiency arises from
the overhead of routing, selection, and aggregation operations in naive sparse
MoE implementations, which can outweigh savings from skipping expert
computation at moderate model and batch sizes.

Overall, our findings demonstrate that Mixture-of-Experts models can generalize
robustly in moderate-scale vision tasks when appropriately regularized, while
exhibiting distinct optimization geometry and routing behavior compared to
dense baselines. At the same time, we find that conditional computation alone is
insufficient to guarantee practical inference speedups in this setting, as
routing and aggregation overheads can dominate at moderate batch sizes and model
scales. Since MoE architectures are designed to increase representational capacity
through expert specialization while bounding per-input computation, their benefits
are expected to become more pronounced in settings with higher data or task
heterogeneity.

\bibliographystyle{IEEEtran}
\bibliography{references}

@article{ScalingLawsForNeuralLanguageModels,
  title={Scaling Laws for Neural Language Models},
  author={Kaplan, Jared and McCandlish, Sam and Henighan, Tom and Brown, Tom B. and Chess, Benjamin and Child, Rewon and Gray, Scott and Radford, Alec and Wu, Jeffrey and Amodei, Dario},
  journal={arXiv preprint arXiv:2001.08361},
  year={2020}
}

@article{AeRviewOfConvolutionalNeuralNetworksInComputerVision,
  title={A review of convolutional neural networks in computer vision},
  author={Zhao, Xia and Wang, Limin and Zhang, Yufei and Han, Xuming and Deveci, Muhammet and Parmar, Milan},
  journal={Artificial Intelligence Review},
  volume={57},
  number={4},
  pages={99},
  year={2024},
  doi={10.1007/s10462-024-10721-6},
  url={https://doi.org/10.1007/s10462-024-10721-6}
}

@article{LargeLanguageModelsASurvey,
  title={Large Language Models: A Survey},
  author={Minaee, Shervin and Mikolov, Tomas and Nikzad, Narjes and Chenaghlu, Meysam and Socher, Richard and Amatriain, Xavier and Gao, Jianfeng},
  journal={arXiv preprint arXiv:2402.06196},
  year={2024},
  url={https://arxiv.org/abs/2402.06196}
}

@article{DeepSpeech2,
  author       = {Dario Amodei and
                  Rishita Anubhai and
                  Eric Battenberg and
                  Carl Case and
                  Jared Casper and
                  Bryan Catanzaro and
                  Jingdong Chen and
                  Mike Chrzanowski and
                  Adam Coates and
                  Greg Diamos and
                  Erich Elsen and
                  Jesse H. Engel and
                  Linxi Fan and
                  Christopher Fougner and
                  Tony Han and
                  Awni Y. Hannun and
                  Billy Jun and
                  Patrick LeGresley and
                  Libby Lin and
                  Sharan Narang and
                  Andrew Y. Ng and
                  Sherjil Ozair and
                  Ryan Prenger and
                  Jonathan Raiman and
                  Sanjeev Satheesh and
                  David Seetapun and
                  Shubho Sengupta and
                  Yi Wang and
                  Zhiqian Wang and
                  Chong Wang and
                  Bo Xiao and
                  Dani Yogatama and
                  Jun Zhan and
                  Zhenyao Zhu},
  title        = {Deep Speech 2: End-to-End Speech Recognition in English and Mandarin},
  journal      = {CoRR},
  volume       = {abs/1512.02595},
  year         = {2015},
  url          = {http://arxiv.org/abs/1512.02595},
  eprinttype    = {arXiv},
  eprint       = {1512.02595},
  bibsource    = {dblp computer science bibliography, https://dblp.org}
}

@misc{deepseekv3technicalreport,
  title={DeepSeek-V3 Technical Report},
  author={{DeepSeek-AI}},
  year={2024},
  eprint={2412.19437},
  archivePrefix={arXiv},
  primaryClass={cs.CL},
  url={https://arxiv.org/abs/2412.19437},
  note={arXiv:2412.19437}
}

@article{shazeer2017outrageously,
  title={Outrageously Large Neural Networks: The Sparsely-Gated Mixture-of-Experts Layer},
  author={Shazeer, Noam and Mirhoseini, Azalia and Maziarz, Krzysztof and Davis, Andy and Le, Quoc and Hinton, Geoffrey and Dean, Jeff},
  journal={arXiv preprint arXiv:1701.06538},
  year={2017},
  url={https://arxiv.org/abs/1701.06538}
}

@article{SwitchTransformers,
  title={Switch Transformers: Scaling to Trillion Parameter Models with Simple and Efficient Sparsity},
  author={Fedus, William and Zoph, Barret and Shazeer, Noam},
  journal={Journal of Machine Learning Research},
  volume={23},
  pages={1--40},
  year={2022},
  url={http://jmlr.org/papers/v23/21-0998.html},
  note={Submitted 8/21; Published 4/22}
}

@article{jacobsoriginalMoE,
  title={Adaptive mixtures of local experts},
  author={Jacobs, Robert A. and Jordan, Michael I. and Nowlan, Steven J. and Hinton, Geoffrey E.},
  journal={Neural Computation},
  volume={3},
  number={1},
  pages={79--87},
  year={1991},
  publisher={MIT Press},
  doi={10.1162/neco.1991.3.1.79}
}

@inproceedings{onlargebatchtraining,
  title={On Large-Batch Training for Deep Learning: Generalization Gap and Sharp Minima},
  author={Keskar, Nitish Shirish and Mudigere, Dheevatsa and Nocedal, Jorge and Smelyanskiy, Mikhail and Tang, Ping Tak Peter},
  booktitle={International Conference on Learning Representations},
  year={2017}
}

@article{krizhevsky2009learning,
  title={Learning multiple layers of features from tiny images},
  author={Krizhevsky, Alex and Hinton, Geoffrey and others},
  year={2009},
  publisher={Toronto, ON, Canada}
}

\end{document}